\documentclass[conference]{IEEEtran}
\IEEEoverridecommandlockouts

\usepackage{cite}
\usepackage{amsmath,amssymb,amsfonts}
\usepackage{algorithmic}
\usepackage{graphicx}
\usepackage{textcomp}
\usepackage{xcolor}
\usepackage{graphics} 
\usepackage{epsfig} 
\usepackage{mathptmx} 
\usepackage{times} 
\usepackage{amsmath} 
\usepackage{amssymb}  
\usepackage{booktabs} 
\usepackage{array}
\usepackage{tabularx}

\def\BibTeX{{\rm B\kern-.05em{\sc i\kern-.025em b}\kern-.08em
    T\kern-.1667em\lower.7ex\hbox{E}\kern-.125emX}}
\begin{document}

\title{Supervised Contrastive Learning for Ordinal Engagement Measurement}

\author{
    Sadaf Safa$^{1,2}$, Ali Abedi$^{1,2}$, Shehroz S. Khan$^{1,2,3}$ \\
    \textit{$^1$KITE Research Institute, University Health Network, Toronto, Canada} \\
    \textit{$^2$Institute of Biomedical Engineering, University of Toronto, Toronto, Canada} \\
    \textit{$^3$College of Engineering and Technology, American University of the Middle East, Kuwait} \\
    \{sadaf.safa, ali.abedi, shehroz.khan\}@uhn.ca
}

\maketitle
\begin{abstract}
Student engagement plays a crucial role in the successful delivery of educational programs. Automated engagement measurement helps instructors monitor student participation, identify disengagement, and adapt their teaching strategies to enhance learning outcomes effectively. This paper identifies two key challenges in this problem: class imbalance and incorporating order into engagement levels rather than treating it as mere categories. Then, a novel approach to video-based student engagement measurement in virtual learning environments is proposed that utilizes supervised contrastive learning for ordinal classification of engagement. Various affective and behavioral features are extracted from video samples and utilized to train ordinal classifiers within a supervised contrastive learning framework (with a sequential classifier as the encoder). A key step involves the application of diverse time-series data augmentation techniques to these feature vectors, enhancing model training. The effectiveness of the proposed method was evaluated using a publicly available dataset for engagement measurement, DAiSEE, containing videos of students who participated in virtual learning programs. The results demonstrate the robust ability of the proposed method for the classification of the engagement level. This approach promises a significant contribution to understanding and enhancing student engagement in virtual learning environments.
\end{abstract}

\begin{IEEEkeywords}
Student Engagement, Supervised Contrastive Learning, Ordinal Classification, Imbalanced Classification.
\end{IEEEkeywords}

\section{Introduction}
\label{sec:introduction}

With the shift towards virtual learning in an increasingly digital world, examining student engagement levels in this new educational setting is crucial \cite{xie2020covid,dickinson2022perceptions}. Recent evidence suggests that students in virtual education exhibit lower engagement than their in-person counterparts \cite{hollister2022engagement}. Understanding and evaluating student engagement in virtual learning environments is essential due to its strong correlation with academic performance \cite{dogan2015student}. Moreover, teachers who are aware of student engagement levels can refine their instructional methods to achieve better outcomes \cite{thomas2022automatic}. However, assessing engagement in virtual settings is challenging, especially in large student groups \cite{Khan2022InconsistenciesReview}. Consequently, there is a clear need to develop automated approaches to measure student participation \cite{karimah2022automatic}. Such methods can help instructors intervene and address engagement concerns more effectively to enhance learning. To tackle this challenge, various machine-learning and deep-learning algorithms have been developed, primarily focusing on analyzing student videos in virtual learning sessions to infer their engagement levels \cite{karimah2022automatic,Khan2022InconsistenciesReview}.

Various researchers have studied student engagement from different perspectives. Sinatra et al. \cite{sinatra2015challenges} presented a person-oriented perspective on student engagement, defining it as the combination of a student's affective, behavioral, and cognitive states at the time of learning. They argued that these states are most effectively captured through fine-grained physiological and behavioral measures, such as facial expressions, facial action units, and body postures. Extending this idea, D'Mello et al. \cite{d2017advanced} suggested that student engagement is dynamic and not fixed over time. They propose that it should be assessed on a fine-grained timescale that ranges from seconds to a few minutes. As such, the task of measuring engagement is framed as a spatiotemporal data analysis problem. In order to gauge engagement levels from video data, it is crucial not only to assess the student's state in each individual video frame but also to analyze changes in the student's states across consecutive frames \cite{abedi2021improving,abedi2023detecting,abedi2021affect,Khan2022InconsistenciesReview,karimah2022automatic}. Machine learning and deep learning techniques applied to videos for engagement measurement can be categorized into two main approaches: end-to-end methods, in which raw video data are directly analyzed by deep learning algorithms, and feature-based methods, where handcrafted features are initially extracted from the raw video and subsequently analyzed using machine learning / deep learning algorithms to output the engagement level of the person in the video \cite{Khan2022InconsistenciesReview,karimah2022automatic}. Both approaches struggle with two main challenges: class confusion and imbalanced distribution of samples across engagement levels. The first challenge is the issue of class confusion \cite{dave2022tclr}, where there's significant overlap or similarity between samples from distinct classes or engagement levels, making them difficult to differentiate. The second challenge is the imbalance in the distribution of samples across different classes, with some levels of engagement containing significantly more samples than others \cite{karimah2022automatic,Khan2022InconsistenciesReview,abedi2021improving,dresvyanskiy2021deep}.

Various strategies have been proposed to deal with the aforementioned challenges and to carry out video-based engagement measurements \cite{karimah2022automatic,Khan2022InconsistenciesReview}. Some of these include sophisticated end-to-end deep learning architectures mostly equipped with attention mechanisms \cite{liao2021deep,ai2022class,huang2019fine}. Other approaches include extracting affective and behavioral features related to engagement from video data and analyzing them using sequential deep-learning models \cite{abedi2021affect,whitehill2014faces,booth2017toward,kaur2018prediction,niu2018automatic,fedotov2018multimodal,thomas2018predicting,huang2019fine,chen2019faceengage,wu2020advanced,kaur2018prediction,ma2021automatic,copur2022engagement,kaur2018prediction}. Techniques such as the Synthetic Minority Oversampling Technique (SMOTE)  \cite{fedotov2018multimodal, Mandia2024AutomaticMethods} and weighted loss functions \cite{abedi2021improving} are also employed to address the issue of imbalanced distribution of samples across different classes.

Contrastive learning has emerged as a powerful tool for learning effective representations from various data modalities, including visual data \cite{chen2020simple,khan2022supervised,khosla2020supervised} as well as multivariate time series data \cite{luo2023time}, and solving classification \cite{chen2020simple,khan2022supervised,khosla2020supervised} and regression \cite{zha2022supervised} problems in various applications. Contrastive learning has demonstrated excellent results in both self-supervised \cite{chen2020simple} and supervised learning settings \cite{khosla2020supervised,khan2022supervised,zha2022supervised}. A deep neural network in a contrastive learning setting usually comprises a convolutional subnetwork (called an encoder) followed by a fully-connected subnetwork (called a projection head). This network is trained to learn effective representations from input data to differentiate between samples in different classes \cite{chen2020simple,khan2022supervised,khosla2020supervised}. Contrastive learning has shown to be effective in handling the aforementioned challenges of class confusion \cite{dave2022tclr,suresh2021not} and imbalanced distribution of samples across classes \cite{jiang2021improving,jiang2021improving}. 
Previous work has also suggested the quantification and annotation of emotions (as combinations of valence and arousal \cite{russell1980circumplex, abedi2021affect}) as ordinal variables to represent a natural order between engagement levels and not merely as distinct categories. We combine these ideas and formulate a supervised contrastive learning approach for ordinal classification of student engagement in online setting. It is expected that this approach should handle the class imbalance and improve overall classification performance between different engagement levels.
Our primary contributions are as follows:
\begin{itemize}
    \item For the first time in the field of engagement measurement \cite{karimah2022automatic,Khan2022InconsistenciesReview,dewan2019engagement}, supervised contrastive learning equipped with augmentation techniques for time series / sequential data \cite{luo2023time,iwana2021empirical} is used to handle class confusion and imbalanced distribution of samples across classes.
    \item Supervised contrastive learning is implemented in an ordinal classification setting \cite{abedi2021affect,frank2001simple} to improve the classification results. 
\end{itemize}

The remainder of this paper is structured as follows: Section \ref{sec:related_work} provides a review of 
existing methods for video-based engagement measurement. In Section \ref{sec:method}, the proposed method is detailed, discussing the techniques and strategies employed. The experimental setup, dataset, descriptions, and experimental results are presented in Section \ref{sec:results}, while Section \ref{sec:conclusion} concludes the paper with future research directions.

\begin{table*}[!ht]
\centering
\small 
\setlength{\tabcolsep}{12pt} 
\renewcommand{\arraystretch}{1} 
\caption{An overview of features, models, and strategies used to deal with class confusion and class imbalance in the related works.}
\label{tab:related_work}
\begin{tabular}{p{2cm}|p{4.5cm}|p{3.5cm}|p{4.5cm}} 
\toprule
\textbf{Ref., Year} & \textbf{Features} & \textbf{Model} & \textbf{Class Imbalance Method}  \\
\midrule
\cite{gupta2016daisee}, 2016 & None (end-to-end) & 3D-CNN and LRCN & None \\
\cite{geng2019learning}, 2019 & None (end-to-end) & 3D CNN & None\\ 
\cite{zhang2019novel}, 2019 & None (end-to-end) & Inflated 3D-CNN & Weighted Loss Function\\ 
\cite{liao2021deep}, 2021 & None (end-to-end) & ResNet-50 with LSTM & Attention Mechanism\\
\cite{abedi2021improving}, 2021 & None (end-to-end) & ResNet with TCN & Weighted Loss Function\\
\cite{hu2022optimized}, 2022 & None (end-to-end) & ShuffleNet & None\\
\cite{ai2022class}, 2022 & None (end-to-end) & Video Transformer & Attention Mechanism\\
\cite{mehta2022threedimensional}, 2022 & None (end-to-end) & 3D DenseNet & Attention Mechanism\\
\cite{selim2022students}, 2022 & None (end-to-end) & EfficientNet and TCN & Attention Mechanism\\
\cite{whitehill2014faces}, 2014 & Box and Gabor filter & SVM & None \\
\cite{booth2017toward}, 2017 & FAU, Optical Flow, and Head Pose & SVM & None \\
\cite{kaur2018prediction}, 2018 & LBP-TOP & Fully-connect Network & None \\
\cite{fedotov2018multimodal}, 2018 & Body Pose, Facial Embedding, Eye Gaze, Speech & Logistic Regression & SMOTE \\
\cite{thomas2018predicting}, 2018 & Gaze, Head Pose, AU & TCN & None \\
\cite{chen2019faceengage}, 2019 & Gaze, Blink, Head Pose, Embedding & RNN & None \\
\cite{wu2020advanced}, 2020 & Gaze, Head pose, Body Pose, C3D & LSTM, GRU & None \\
\cite{ma2021automatic}, 2021 & Gaze, Head Pose, FAU, and C3D & Neural Turing Machines & None \\
\cite{copur2022engagement}, 2022 & Eye Gaze, Head Pose, and FAU & LSTM & None \\
\cite{abedi2021affect}, 2021 & Valence, Arousal, Blink, and Hand Pose & Ordinal TCN and LSTM & None \\
\cite{thomas2022automatic}, 2022 & FAU, Micro-Macro-Motion & TCN & None \\
\cite{abedi2023detecting}, 2023 & Valence, Arousal, Blink, and Hand Pose & TCN Autoencoder & None \\
\cite{singh2023have}, 2023 & FAU, Eye Gaze, and Head Pose & TCN and LSTM & None \\
\cite{shiri2023recognition}, 2023 & None (end-to-end) & ConvNeXtLarge + ensemble GRU and Bi-GRU & None \\
\cite{li2024multimodal}, 2024 & Visual, Textual, Acoustic  & Haar-MGL (Multimodal Graph Learning) & None  \\
\cite{geng2024engagement}, 2024 & None (end-to-end) & SFRNet (ResNet-18 + ModernTCN) & None\\
\cite{liu2024msc}, 2024 & Eye State, Head posture, Writing Status, Mobile Phone Usage& MSC-Trans (Multi-feature-fusion Network) & None \\
\cite{shiri2024detection}, 2024 & None (end-to-end) & Hybrid models (EfficientNetV2-L + LSTM/GRU/Bi-LSTM/Bi-GRU) & None  \\
\cite{das2024enhancing}, 2024 & Facial landmarks, Head Pose, Gaze, Hand Pose, Valence, Arousal& SVM, kNN, and Random Forest & None \\
\cite{Mandia2024AutomaticMethods}, 2024 & Facial Features & Begging Ensemble Hybrid (1D ResNet +1D CNN) & SMOTE  \\
\cite{Zhang2024AExpressions}, 2024 & Facial Regions & Facial Masked Autoencoder (FMAE) & None \\
\cite{10.1007/978-3-031-78201-5_21}, 2024 & Facial landmarks & ST-GCN, Ordinal ST-GCN & None  \\
\textbf{  Proposed} & Valence, Arousal, Blink, Hand Pose & Ordinal TCN and LSTM & Supervised Contrastive Learning \\
\bottomrule
\end{tabular}
\end{table*}

\section{Related Work}
\label{sec:related_work}
Video analysis remains at the forefront of major research advancements in measuring engagement in online learning settings
\cite{karimah2022automatic}. The methodologies employed to address this task, primarily deep learning-based, are broadly categorized into end-to-end and feature-based approaches. \cite{karimah2022automatic, abedi2021affect,abedi2023detecting}. Table \ref{tab:related_work} provides an overview of these methods. End-to-end approaches typically leverage deep learning techniques to input raw video frames to spatiotemporal Convolutional Neural Networks (CNNs), such as 3D CNNs \cite{gupta2016daisee,geng2019learning,zhang2019novel,mehta2022threedimensional}, or a combination of 2D CNNs and sequential neural networks such as Recurrent Neural Networks (RNNs) \cite{liao2021deep}, Temporal Convolutional Networks (TCNs) \cite{abedi2021improving}, and transformers \cite{ai2022class}. On the other hand, feature-based approaches have been designed to decouple feature extraction from sequential modeling. These techniques extract features, either handcrafted behavioral and affective features \cite{booth2017toward,niu2018automatic,fedotov2018multimodal,thomas2018predicting,huang2019fine,chen2019faceengage,wu2020advanced,ma2021automatic,copur2022engagement,abedi2021affect} or traditional computer-vision features \cite{whitehill2014faces,kaur2018prediction}, from consecutive video frames, and subsequently analyze them using sequential models such as RNNs \cite{abedi2021affect,copur2022engagement,wu2020advanced,chen2019faceengage,ma2021automatic,niu2018automatic} and TCNs \cite{abedi2021affect,thomas2018predicting,thomas2022automatic} or non-sequential models such as bag-of-words \cite{abedi2023bag}. A notable advancement in this domain is the use of Spatial-Temporal Graph Convolutional Networks (ST-GCNs) \cite{10.1007/978-3-031-78201-5_21}, which inherently analyze spatial-temporal facial landmarks and learn these features without requiring raw facial videos. Such models reduce the dependency on raw video data by focusing on learned landmark representations. In contrast, some feature-based methodologies bypass sequential analysis by directly computing functionals of the extracted features (e.g., mean, standard deviation), which are then fed into non-sequential models such as support vector machines \cite{booth2017toward,whitehill2014faces,chen2019faceengage}. While effective, these functional-based methods may suffer from information loss when determining the optimal window size for functional calculation \cite{abedi2021affect}.

Although existing techniques for engagement measurement have shown promising results, they still face major challenges, such as class confusion and an imbalanced distribution of samples across engagement classes \cite{abedi2021improving,liao2021deep,fedotov2018multimodal}. Underrepresentation of certain engagement classes can lead the model to develop a bias towards overrepresented classes, thereby compromising prediction accuracy on minority classes \cite{abedi2021improving,liao2021deep,fedotov2018multimodal}. Some efforts have focused on overcoming these issues, such as incorporating attention mechanisms in deep neural networks to better distinguish between different levels of participation and counter-class confusion \cite{liao2021deep,huang2019fine,ai2022class} and applying oversampling techniques such as SMOTE for underrepresented classes \cite{fedotov2018multimodal, Mandia2024AutomaticMethods}, along with weighted loss functions to manage data imbalance \cite{abedi2021improving,liao2021deep,zhang2019novel}. Table 1 summarizes related work by highlighting key features, models, and techniques used to address class confusion and data imbalance.

While contrastive learning techniques have demonstrated considerable promise in various applications \cite{chen2020simple,khan2022supervised,khosla2020supervised,luo2023time}, they have yet to be applied in the realm of engagement measurement \cite{karimah2022automatic,Khan2022InconsistenciesReview,dewan2019engagement}. These methodologies have the capability to address issues of class imbalance \cite{khan2022supervised} and class confusion inherent in engagement measurement problems, offering a potential enhancement in the efficacy of predictive models. This paper introduces the first utilization of supervised contrastive learning in the domain of engagement measurement. The focus is on developing sequential deep-learning models in a supervised contrastive learning setting and subsequently articulating them as an ordinal classification problem to further improve performance.

\section{Method}
\label{sec:method}
\begin{figure*}[!htp]
\begin{center}
\includegraphics[scale=0.3]{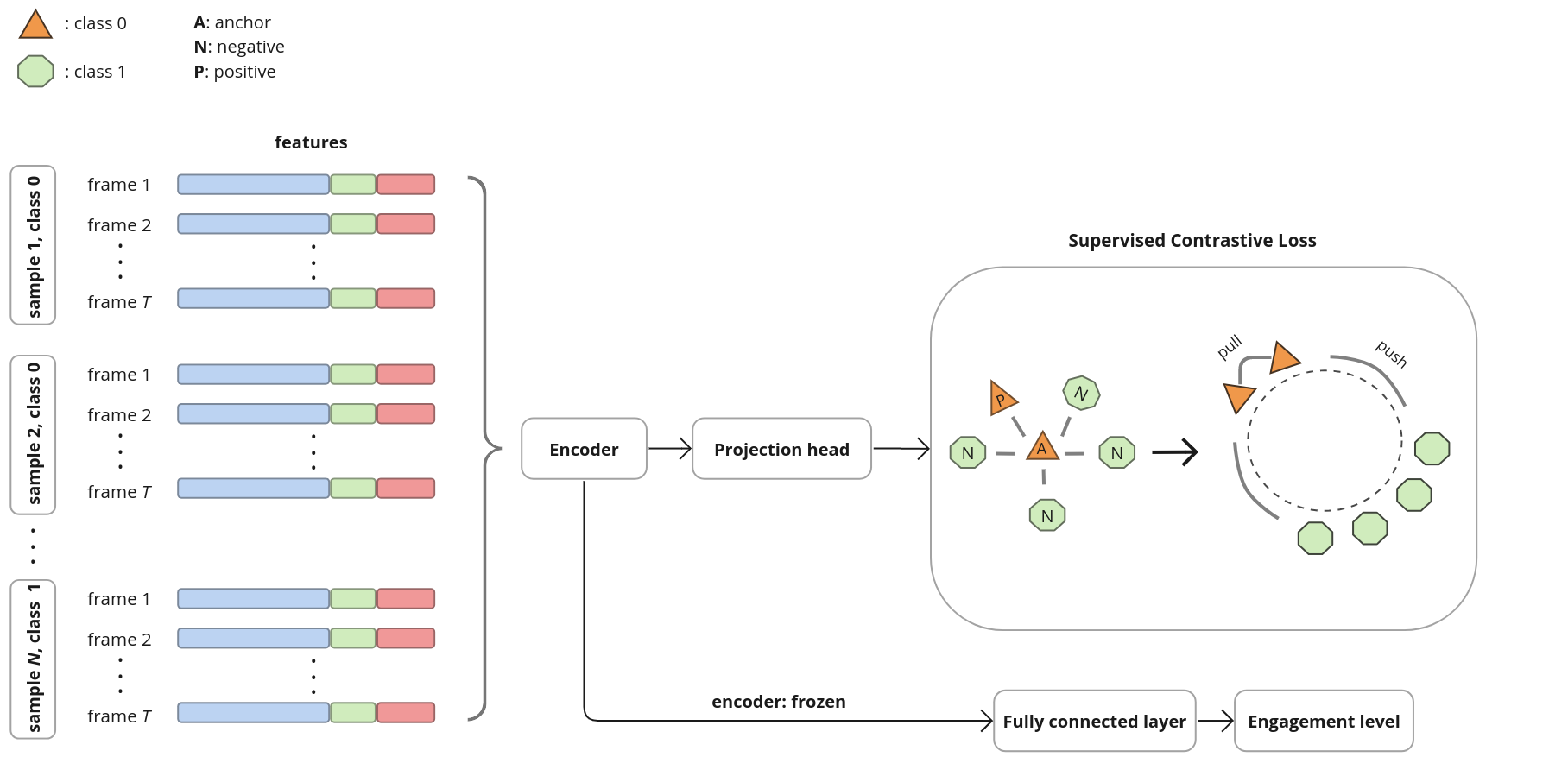}
\caption{Supervised contrastive learning for a binary classification problem.}\label{fig:cl}
\end{center}
\end{figure*}
The proposed method for engagement measurement utilizes a video sample as input, capturing a person sitting in front of a webcam during a virtual learning session. The method outputs a discrete ordinal level representing the individual's engagement in the video. 
. The intermediates steps are as follows:

\noindent
\paragraph{Feature extraction}
In accordance with the definition of engagement in learning settings \cite{sinatra2015challenges,d2017advanced}, affective and behavioral features are extracted from successive video frames \cite{abedi2021affect}. This approach allows constructing a sequence of feature vectors or a multi-variate time series for each input video sample. Affect features include values of valence and arousal (2 features) and a 256-element latent affective feature vector \cite{abedi2021affect}. Twelve behavioral features include eye-closure facial action unit intensity (1 feature) \cite{ranti2020blink,baltrusaitis2018openface}, x and y components of eye gaze direction (2 features), x, y, and z components of head location (3 features), pitch, yaw, and roll as head pose (3 features), x, y, and z components of wrist location (3 features) \cite{baltrusaitis2018openface}. For the purpose of joint fusion, the latent affective feature vector is fed to a trainable auxiliary fully-connected network to reduce its dimensionality to 32, which is further concatenated to valence and arousal as well as behavioral features to construct the feature vector for the input frame. Please refer to section \ref{sub_sec: experimental_setup} for feature extraction details.

\noindent
\paragraph{Supervised Contrastive Learning}
As a built-in first step in supervised contrastive learning \cite{khosla2020supervised}, data augmentation techniques (more details in Section \ref{sub_sec: experimental_setup}) are applied to the original data samples (in our case multivariate time series) \cite{iwana2021empirical,luo2023time}. The union of the original and augmented data samples and their associated labels is used for model training $\{x_i, y_i\}, i \in I = \{1, 2, \ldots, N\}$, which is carried out through two distinct phases, detailed as follows.

In the first phase, a neural network comprising an encoder $enc(.)$ trailed by a projection head $proj(.)$ is trained. While the encoder is a sequential model, such as Long Short-Term Memory (LSTM) and TCN, the projection head is a fully-connected network. The encoder and the projection head analyze input sequences and generate $1$-dimensional vectors called representations $z_i = proj(enc(x_i)), i \in I = \{1, 2, \ldots, N\}$. The representations are input to a supervised contrastive loss function as follows \cite{khosla2020supervised}.

\begin{equation}
    \mathcal{L}=\sum_{i \in I} \frac{-1}{|{P}(i)|}
    \sum_{p \in {P}(i)} \log \frac{\exp \left({z}_{i} \cdot {z}_{p}/\tau\right)}{\sum_{a \in {A}(i)} \exp \left(\boldsymbol{z}_{i} \cdot \boldsymbol{z}_{a}/\tau\right)}
\end{equation}

$A(i)$ is the set of all samples in the training set, excluding $i$, and $P(i)$ is the set of all samples in $A(i)$ that share the same label as $i$. $|P(i)|$ represents the cardinality of $P(i)$. In this context, the term anchor is assigned to the index $i$, while the positive samples are used to denote samples with identical labels, and the negative samples refer to samples that possess different labels. In the course of loss function minimization and representation learning, the numerator within the inner summation is maximized, thereby enhancing the similarity (quantified via the dot product) among representations of samples with the same labels, i.e., positive pairs. Conversely, the denominator is minimized to lessen the similarity between the representations of samples with differing labels, i.e., negative pairs. As a result, the encoder and projection head learn to represent samples (shown as $z_i = proj(enc(x_i))$) such that samples with the same labels are represented similarly, while samples with different labels are represented differently (see Figure \ref{fig:cl}). The temperature parameter $\tau$ balances the model's sensitivity to the similarity between samples \cite{wang2021understanding}.

In the second phase, the projection head is discarded, and the encoder is frozen. Using $\{x_i, y_i\}, i \in I = \{1, 2, \ldots, N\}$, a new fully connected neural network is trained on top of the frozen representations generated by the frozen encoder through the minimization of a cross-entropy loss function.

\noindent
\paragraph{Incorporating Ordinality of the Engagement Levels}
The different levels of engagement (from disengaged to fully engaged) follow a natural order, which is used by researchers to annotate student engagement datasets. \cite{yannakakis2018ordinal, whitehill2014faces}
Despite this, many current machine learning and deep learning methodologies approach engagement measurement as a categorical classification problem \cite{Khan2022InconsistenciesReview,karimah2022automatic}. Building upon Abedi and Khan \cite{abedi2021affect}, who have developed ordinal models for engagement measurement, our approach leverages and integrates the ordinality of the engagement variable, thereby viewing engagement measurement as an ordinal classification problem followed by contrastive learning. In the ordinal framework \cite{frank2001simple}, a $C$-class classification problem is transformed into $(C-1)$ binary classification problems, each differentiating between samples above or below a particular class; the first binary classifier differentiates between samples in class $0$ and those with class labels greater than $0$, the second differentiates between samples in classes $0$ and $1$ and the remaining samples with class labels greater than $1$, and so forth. At inference time for a sample $x_i$, the probability estimates of the $(C-1)$ binary classifiers $f_c(.), c = 0, 1, ..., C-2$ are combined to yield $C$ probability estimates for $C$ classes, thereby facilitating appropriate inferences. The probabilities are calculated as:

\begin{equation}
    f(y_t = c) =
    \begin{cases} 
        1 - f(y_t > 0), & \text{if } c = 0, \\
        f(y_t > c - 1) - f(y_t > c), & \text{if } 0 < c < C - 1, \\
        f(y_t > C - 2), & \text{if } c = C - 1.
    \end{cases}
\end{equation}

\section{Experiments and Results}
\label{sec:results}
This section introduces the dataset used for evaluation, details the experimental setup, and presents the results under different configurations, alongside a comparison with previous methods.

\subsection{DAiSEE Dataset} 
\label{sec:dataset} 
The dataset used in this study is DAiSEE  \cite{gupta2016daisee}, which comprises $9,068$ video samples from $112$ students, each lasting $10$ seconds and recorded at a resolution of 640 × 480 pixels at 30 frames per second.The videos are annotated for four affective states—engagement, boredom, confusion, and frustration—each categorized into four levels of intensity.  The distribution of samples across engagement levels within the train, validation, and test sets is detailed in Table \ref{tab:dataset}. This dataset has four engagement levels, labelled as $0, 1, 2$ and $3$. The distribution of samples for engagement levels $0$ and $1$ across train, validation, and test sets show high imbalance, (e.g., only $4.93\%$ of the test samples have levels $0$ and $1$). 
The dataset's train and validation sets are combined and are used for model training, while the testing was conducted on $1,784$ video samples from the test set. 

\begin{table}[ht]
\small
\caption{Distribution of samples across various engagement levels in the training, validation, and testing subsets of the DAiSEE dataset \cite{gupta2016daisee}.}
\label{tab:dataset}
\centering
\begin{tabular}{p{.2\linewidth}p{.15\linewidth}p{.15\linewidth}p{.15\linewidth}}
\hline
level & train & validation & test\\
\hline
0 & 2& 23 & 4\\
\hline
1 & 213 & 143 & 84\\
\hline
2 & 2617 & 813 & 882\\
\hline
3 & 2494 & 450 & 814\\
\hline
total & 5358 & 1429 & 1784\\
\hline
\end{tabular}
\end{table}


\begin{table*}[ht]
\centering
\caption{Total accuracy (over all samples in all classes) and class-specific precisions and recalls of different settings of the proposed method on the test set of the DAiSEE dataset, using different feature sets, models, and loss functions/training methods, ordinal and regular four-class classification. 
Values highlighted in bold represent the best results.
}
\label{tab:recall-prec-acc-tab}
\resizebox{\textwidth}{!}{%
\begin{tabular}{@{}lllllllllllll@{}}
\toprule
& & & & \multicolumn{2}{c}{Class 0} & \multicolumn{2}{c}{Class 1} & \multicolumn{2}{c}{Class 2} & \multicolumn{2}{c}{Class 3} \\\midrule
Features & Model & Loss Function/Training method & Total Acc. & Prec. & Rec. & Prec. & Rec. & Prec. & Rec. & Prec. & Rec. \\ 
\midrule
Affect + Behavioral & LSTM & (a) Cross-entropy & 0.6110 & 0 & 0 & 0.1034 & 0.0741 & 0.6314 & 0.6860 & 0.6260 & 0.5869 \\ \midrule
Affect + Behavioral & LSTM & (b) Class-weighted cross-entropy & 0.5710 & 0.0270 & 0.2500 & 0.1063 & 0.0610 & 0.5840 & 0.6910 & 0.6180 & 0.4929 \\ \midrule
Affect + Behavioral + Latent & LSTM & (a) Cross-entropy & 0.5925 & 0 & 0 & 0.1147 & 0.0864 & 0.6250 & 0.6410 & 0.5945 & 0.5945 \\ \midrule
Affect + Behavioral + Latent & LSTM & (b) Class-weighted cross-entropy & 0.5746 & 0.0833 & 0.2500 & 0.0577 & 0.0370 & 0.6051 & 0.6350 & 0.5815 & 0.5650 \\ \midrule
Affect + Behavioral + Latent & LSTM & (c) Contrastive + Cross-entropy & 0.6297 & 0 & 0 & 0 & 0 & 0.6347 & 0.7166 & 0.6307 & 0.6023 \\ \midrule
Affect + Behavioral + Latent & LSTM & (d) Contrastive + Class-weighted cross-entropy & 0.6220 & 0 & 0 & 0.1100 & 0.0247 & 0.6280 & 0.7080 & 0.6290 & 0.5910 \\ \midrule
Affect + Behavioral + Latent & LSTM & (e) Contrastive + Cross-entropy (Augmented) & 0.6295 & 0 & 0 & 0.3750 & 0.3214 & 0.6762 & 0.6179 & 0.6109 & 0.6769 \\ \midrule
Affect + Behavioral + Latent & LSTM & (f) Contrastive + Cross-entropy (Augmented) (Ordinal) & 0.6570 & 0 & 0 & 0.3596 & \textbf{0.3810} & 0.6843 & 0.6587 & 0.6608 & \textbf{0.6867} \\ \midrule
Affect + Behavioral & TCN & (a) Cross-entropy & 0.6000 & 0 & 0 & 0.0488 & 0.0246 & 0.6400 & 0.6600 & 0.5950 & 0.6060 \\ \midrule
Affect + Behavioral & TCN & (b) Class-weighted cross-entropy & 0.5711 & 0 & 0 & 0.0256 & 0.0123 & 0.6250 & 0.5570 & 0.5560 & 0.6470 \\ \midrule
Affect + Behavioral & TCN & (c) Contrastive + Cross-entropy & 0.6030 & 0 & 0 & 0.0540 & 0.0247 & 0.6205 & 0.7142 & 0.6107 & 0.5431 \\ \midrule
Affect + Behavioral & TCN & (d) Contrastive + Class-weighted cross-entropy & 0.5827 & 0 & 0 & 0.0106 & 0.0123 & 0.6280 & 0.6710 & 0.6080 & 0.5470 \\ \midrule
Affect + Behavioral + Latent & TCN & (a) Cross-entropy & 0.6050 & 0 & 0 & 0.1315 & 0.0617 & 0.6201 & 0.6780 & 0.6127 & 0.5842 \\ \midrule
Affect + Behavioral + Latent & TCN & (a’) Cross-entropy (Augmented) (Ordinal) & \textbf{0.6754} & 0 & 0 & 0.3617 & 0.2024 & 0.6667 & \textbf{0.7574} & \textbf{0.7123} & 0.6388 \\ \midrule
Affect + Behavioral + Latent & TCN & (b) Class-weighted cross-entropy & 0.5870 & 0 & 0 & 0.0106 & 0.0123 & 0.6283 & 0.67131 & 0.6080 & 0.5470 \\\midrule
Affect + Behavioral + Latent & TCN & (c) Contrastive + Cross-entropy & 0.5990 & \textbf{0.5000} & \textbf{0.2500} & 0 & 0 & 0.6110 & 0.6990 & 0.6050 & 0.5520 \\ \midrule
Affect + Behavioral + Latent & TCN & (d) Contrastive + Class-weighted cross-entropy & 0.5873 & 0 & 0 & 0.2090 & 0.0740 & 0.6160 & 0.6480 & 0.5890 & 0.5760 \\ \midrule
Affect + Behavioral + Latent & TCN & (e) Contrastive + Cross-entropy (Augmented) & 0.6547 & 0 & 0 & 0.4386 & 0.2976 & 0.6693 & 0.6769 & 0.6570 & 0.6708 \\ \midrule
Affect + Behavioral + Latent & TCN & (f) Contrastive + Cross-entropy (Augmented) (Ordinal) & 0.6732 & 0 & 0 & \textbf{0.4754} & 0.3452 & \textbf{0.6902} & 0.6995 & 0.6711 & 0.6818 \\
\bottomrule
\end{tabular}}
\end{table*}

\subsection{Experimental Setup}
\label{sub_sec: experimental_setup}
To extract the behavioral features (see Section \ref{sec:method}), the pre-trained OpenFace \cite{baltrusaitis2018openface} was utilized. Additionally, the pre-trained Emotion Face Alignment Network (EmoFAN) \cite{toisoul2021estimation} on the Affect from the InterNet (AffectNet) dataset \cite{mollahosseini2017affectnet} was used to extract the affect features and latent affective features.

In the supervised contrastive learning and ordinal framework (see Section \ref{sec:method}), the auxiliary fully-connected network responsible for the joint fusion of the latent affective feature vectors was trained in the first phase alongside the encoder and projection head. This network consisted of two layers of sizes $256 \times 128$ and $128 \times 32$. The performance of two sequential models was evaluated to serve as the encoder. In Long Short Term Memory (LSTM), the best results were obtained with two layers with 256 hidden units. The optimal parameters for Temporal Convolution Network (TCN) were $8, 256, 16,$ and $0.1$ for the number of levels, hidden units, kernel size, and dropout, respectively. The output of the final hidden state of the encoder was fed to a fully connected layer with $256 \times 128$ neurons as the projection head. In the second phase described in Section \ref{sec:method}, two fully connected layers with $256 \times 128$ and $128 \times num_{classes}$ neurons are used for inference. For non-ordinal four-class classification, $num_{classes}$ was $4$, the activation function was Softmax, and the loss function was cross-entropy loss. Furthermore, a performance evaluation of the weighted cross-entropy loss was conducted, where the weights were determined inversely proportional to the percentage of samples in the dataset \cite{zhang2019novel, jiang2021improving}. For ordinal classification, where the four-class classification problem is divided into three binary classification problems, $num_{classes}$, activation function, and the loss function for each binary classifiers were $1$, sigmoid, and binary cross-entropy loss, respectively.

Time-series augmentation techniques were applied to the sequences of feature vectors including jittering, magnitude scaling, time shifting, permutation, and flipping \cite{iwana2021empirical}. Jittering is achieved by adding Gaussian noise with a mean of zero and a standard deviation of one-tenth of the peak-to-peak value of the feature over consecutive frames. Magnitude scaling was applied randomly with a relative scale of 0.75. Time shifting was implemented with five units at random. These parameters were selected in such a way as not to change the nature of the signal in terms of engagement level. These techniques were applied randomly to the samples in the training and validation sets of the DAiSEE dataset to increase the total number of samples in the minority classes of 0 and 1 by a factor of ten, and the majority classes of 2 and 3 by a factor of one and a half, giving the most optimal result. No data augmentation was applied to the test set to avoid information leakage. 

The evaluation metrics were accuracy, class-specific precision and recall, and confusion matrix. The experiments were implemented in PyTorch \cite{paszke2019pytorch} and scikit-learn \cite{pedregosa2011scikit} on a  server with  64  GB  of  RAM  and  NVIDIA Tesla P100 PCIe  12  GB  GPU.

\subsection{Results}
\label{sec:experimental_results}
\begin{table*}[ht]
\caption{The confusion matrices on the test set of DAiSEE dataset for previous (a) 
ResNet + TCN \cite{abedi2021improving} and (b) affective and behavioral features + Ordinal TCN \cite{abedi2021affect} approaches, and two settings of the proposed method, ordinal four-class classification using contrastive loss with data augmentation using (c) LSTM and (d) TCN. 
Values highlighted in bold represent the best results.}
\centering
\includegraphics[scale=1.]{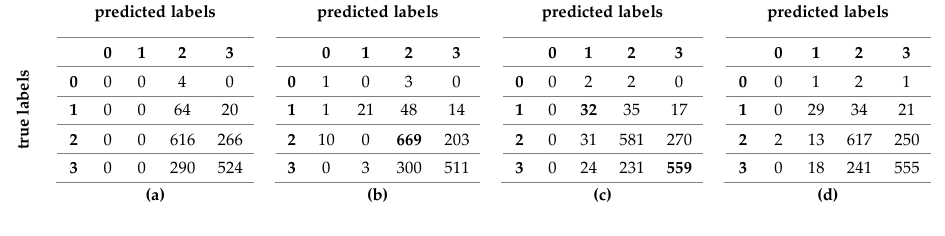}
\label{tab:confusion_matrices}
\end{table*}

Table \ref{tab:recall-prec-acc-tab} outlines an ablation study for the proposed method, including total accuracy over all samples in all classes and class-specific precisions and recalls of different settings of the proposed method on the test set of the DAiSEE dataset \cite{gupta2016daisee}, using different feature sets, models, and loss functions / training methods. Table \ref{tab:recall-prec-acc-tab} shows that the model's performance is notably superior when utilizing all features compared to excluding latent affective features. This highlights the importance of jointly fusing latent affective features, employing an auxiliary network, and concatenating them with affect and behavioral features to construct feature vectors. 

\begin{table}[ht]
\small  
\caption{Engagement level classification accuracy on the test set of the DAiSEE dataset using the best setting of the proposed method and the previous methods.}
\centering
\renewcommand{\arraystretch}{1.2}  
\begin{tabularx}{\columnwidth}{Xl}
\toprule
\textbf{Method} & \textbf{Accuracy} \\
\midrule
3D CNN \cite{gupta2016daisee} & 0.4810 \\
Inflated 3D-CNN \cite{zhang2019novel} & 0.5240 \\
behavioral features + Support Vector Machine & 0.5575 \\
3D-CNN + LSTM \cite{abedi2021improving} & 0.5660 \\
Long-term Recurrent CNN \cite{gupta2016daisee} & 0.5790 \\
Hybrid R(2+1)D and spatio-temporal block \cite{Sathisha2024FacialDiscrimination} & 0.5862\\
Functional features + Random Forest  & 0.5909 \\
ResNet-50 with LSTM with Attention \cite{liao2021deep} & 0.5880 \\
3D-CNN + TCN \cite{abedi2021improving} & 0.5990 \\
behavioral features + LSTM with attention \cite{huang2019fine} & 0.6000 \\
2D ResNet + LSTM \cite{abedi2021improving} & 0.6150 \\
behavioral features + Neural Turing Machine \cite{ma2021automatic} & 0.6130 \\
3D DenseNet with Attention \cite{mehta2022threedimensional} & 0.6360 \\
ResNet + TCN \cite{abedi2021improving} & 0.6390 \\
ShuffleNet \cite{hu2022optimized} & 0.6390 \\
PANet + STformer \cite{Su2024LeveragingDetection} & 0.6400\\
EfficientNet + TCN \cite{selim2022students} & 0.6467 \\
Self-Supervised FMAE \cite{Zhang2024AExpressions} & 0.6474 \\
ViT-TCN (pre-trained) \cite{Zhang2024AExpressions} & 0.6558\\
affective and behavioral features + Ordinal TCN \cite{abedi2021affect} & 0.6740 \\
EfficientNet + LSTM \cite{selim2022students} & 0.6748 \\

\textbf{Proposed} & 0.6732 \\
\bottomrule
\end{tabularx}
\label{tab:comparison}
\end{table}

The overall superiority of TCN over LSTM is attributed to TCN's ability to handle long sequences and mitigate issues related to exploding / vanishing gradients. However, it is worth noting that for classes $1$ and $3$, LSTM, row (f) in Table \ref{tab:recall-prec-acc-tab}, outperformed TCN in terms of recall. Across different combinations of features and models, contrastive learning consistently outperformed traditional learning with cross-entropy loss, as evident from rows (a) and (c). The results deteriorate when weighted cross-entropy is applied to both contrastive and traditional learning, as seen in rows (a) and (b), and rows (c) and (d). This is due to a severe imbalance in classes $0$ and $1$ in the dataset. Incorporating data augmentation in contrastive learning and training models in the ordinal framework significantly improved the results, as observed in rows (c) and (e), and (e) and (f). This underscores the importance of appropriate time series augmentation techniques in contrastive learning and incorporating engagement ordinality in model design. Although row ($\text{a}^{\prime}$) with cross-entropy loss function and augmentation in the ordinal setting achieves the highest accuracy, the contrastive learning approach with augmentation in the ordinal setting (row (f)) achieves the highest precision and recall in classes $1$ and $2$ alongside a competitive accuracy. This indicates the success of the proposed method in addressing the class imbalance problem and reducing class confusion.

Table \ref{tab:confusion_matrices} shows the confusion matrices of (a) one of the previous end-to-end approaches ResNet + TCN \cite{abedi2021improving}, (b) one of the previous feature-based approaches \cite{abedi2021affect}, and two settings of the proposed method, ordinal four-class classification using contrastive loss with data augmentation using (c) LSTM and (d) TCN as the sequential model on the test set of the DAiSEE dataset. From Table \ref{tab:confusion_matrices}, the proposed method in (c) is superior to the previous methods in the classification of the samples in classes 1 and 3, showing its superiority in handling class imbalance and class confusion in the problem of engagement measurement. However, when it comes to the classification of class 2, TCN in (d) is superior to LSTM in (c).

Table \ref{tab:comparison} provides comparisons of engagement level classification accuracy on the test set of the DAiSEE dataset using the previous methods and the best setting of the proposed method. The proposed method surpasses the majority of previous approaches, except for marginal improvement of EfficientNet + LSTM \cite{selim2022students} and affective and behavioral features + Ordinal TCN \cite{abedi2021affect}. Nevertheless, it outperforms all other prior methods, most of which utilized specialized techniques, such as attention mechanisms \cite{liao2021deep,mehta2022threedimensional} or techniques such as weighted loss functions \cite{zhang2019novel,abedi2021improving} to handle class confusion and class imbalance challenges in the engagement measurement problem.

\section{Conclusion}
\label{sec:conclusion}
This paper introduced a novel approach for engagement measurement from videos, specifically addressing class confusion and class imbalance challenges in the field. Utilizing supervised contrastive learning with augmentation techniques in an ordinal setting, sequential models were successfully trained for feature-based engagement measurement, demonstrating their effectiveness on a publicly available engagement measurement video dataset. Moving forward, we aim to extend our evaluations to other student engagement measurement datasets containing other modalities (e.g., speech) and explore other applications, such as patient engagement in virtual rehabilitation. Future research endeavors will involve incorporating audio data and contextual information, such as the progress and barriers in educational and rehabilitation programs, to enable comprehensive engagement measurement, encompassing cognitive and context-oriented engagement assessments.

\bibliographystyle{IEEEtran}
\bibliography{ref,ref-m}


\end{document}